\pdfoutput=1
\documentclass[10pt,logo,copyright]{nvidiatechreport}
\usepackage[authoryear,sort&compress,round]{natbib}

\usepackage[utf8]{inputenc} 
\usepackage[T1]{fontenc}    

\usepackage{parskip}        
\usepackage{url}            
\usepackage{hyperref}       
\usepackage{booktabs}       
\usepackage{amsfonts}       
\usepackage{nicefrac}       
\usepackage{microtype}      
\usepackage{xcolor}         
\usepackage[dvipsnames]{xcolor} 
\usepackage{graphicx}
\usepackage{animate}        
\usepackage{subcaption}
\usepackage{tabularx}
\usepackage{makecell}
\usepackage{adjustbox}
\usepackage{setspace}
\newcolumntype{M}[1]{>{\centering\arraybackslash}m{#1}}
\usepackage{float}
\usepackage{tikz}
\usetikzlibrary{positioning,shapes,arrows}
\usepackage{amsmath,amsfonts,bm, bbm,leftindex}
\usepackage{multirow}
\usepackage{comment}
\usepackage{gensymb}
\usepackage{lipsum}
\usetikzlibrary{arrows.meta, positioning, fit}
\usepackage[para]{threeparttable}
\usepackage{tikz}
\usetikzlibrary{tikzmark}

\definecolor{darkred}{rgb}{0.7, 0.0, 0.0}

\usepackage{pifont}
\newcommand{\cmark}{\ding{51}} 
\newcommand{\xmark}{\ding{55}} 
\usepackage{wrapfig}

\usepackage[nameinlink]{cleveref}
\crefname{equation}{Eq.}{Eqs.}
\crefname{figure}{Fig.}{Figs.}
\crefname{section}{Sec.}{Sec.}
\crefname{appendix}{App.}{App.}
\crefname{table}{Tab.}{Tabs.}
\crefname{algorithm}{Algo}{Algo}
\crefname{thm}{Thm}{Thm}
\Crefname{thm}{Thm}{Thm}
\crefname{prop}{Prop}{Prop}
\usepackage{longtable}
\usepackage{mathtools}
\usepackage{siunitx}
\usepackage{booktabs}
\usepackage[table,xcdraw]{xcolor}

\newcommand{\Exp}[2]{\mathbb{E}_{\,#1}\!\left[\,#2\,\right]}
\newcommand{\method}{DoorMan}

\newcommand{\minisection}[1]{\noindent{\textbf{#1}.}}

\usepackage{booktabs, tabularx, pifont, array, bbm, float, makecell, multirow, }
\usepackage{ragged2e}
\definecolor{nvidiagreen}{HTML}{76B900}
\definecolor{bestrow}{HTML}{E1EBD7}
\newcolumntype{Y}{>{\raggedleft\arraybackslash}X}
\newcolumntype{L}[1]{>{\raggedright\arraybackslash}p{#1}}

\usepackage{amsfonts}
\usepackage{bm}
\usepackage{amsmath}
\usepackage{mathtools}
\usepackage{multirow}
\usepackage{booktabs}
\usepackage{caption}
\usepackage{enumitem}
\usepackage{wrapfig,lipsum,booktabs}
\usepackage{caption}
\usepackage{bbm}
\usepackage{multirow}
\usepackage{pifont}
\usepackage[linesnumbered,ruled,vlined]{algorithm2e}

\SetCommentSty{mycommfont}
\SetAlCapFnt{\small} 
\usepackage{etoc}
\SetAlgorithmName{Algo}{Algo}{}

\renewcommand{\paragraph}[1]{{\vspace{1mm}\noindent \bf #1}.}

\newcommand{\crefnames}[3]{%
  \@for\next:=#1\do{%
    \expandafter\crefname\expandafter{\next}{#2}{#3}%
  }%
}

\title{Opening the Sim-to-Real Door for Humanoid Pixel-to-Action Policy Transfer}

\begin{document}

\author{
  Haoru Xue$^{1,2*}$,
  Tairan He$^{1,3*}$,
  Zi Wang$^{1*}$,
  Qingwei Ben$^{1,4}$,
  Wenli Xiao$^{1,3}$,
  Zhengyi Luo$^{1}$,
  Xingye Da$^{1}$,
  Fernando Casta\~neda$^{1}$,
  Guanya Shi$^{3}$,
  Shankar Sastry$^{2}$,
  Linxi ``Jim'' Fan$^{1\dagger}$,
  Yuke Zhu$^{1\dagger}$\\
  \small $^{1}$NVIDIA \quad
         $^{2}$UC Berkeley \quad
         $^{3}$CMU \quad
         $^{4}$CUHK\\
  \small $^{*}$Equal Contribution \quad
         $^{\dagger}$Project Leads
}

    

\makeatletter
\let\oldmaketitle\maketitle
\renewcommand{\maketitle}{%
  \oldmaketitle%
  \begin{center}
    \includegraphics[width=\textwidth]{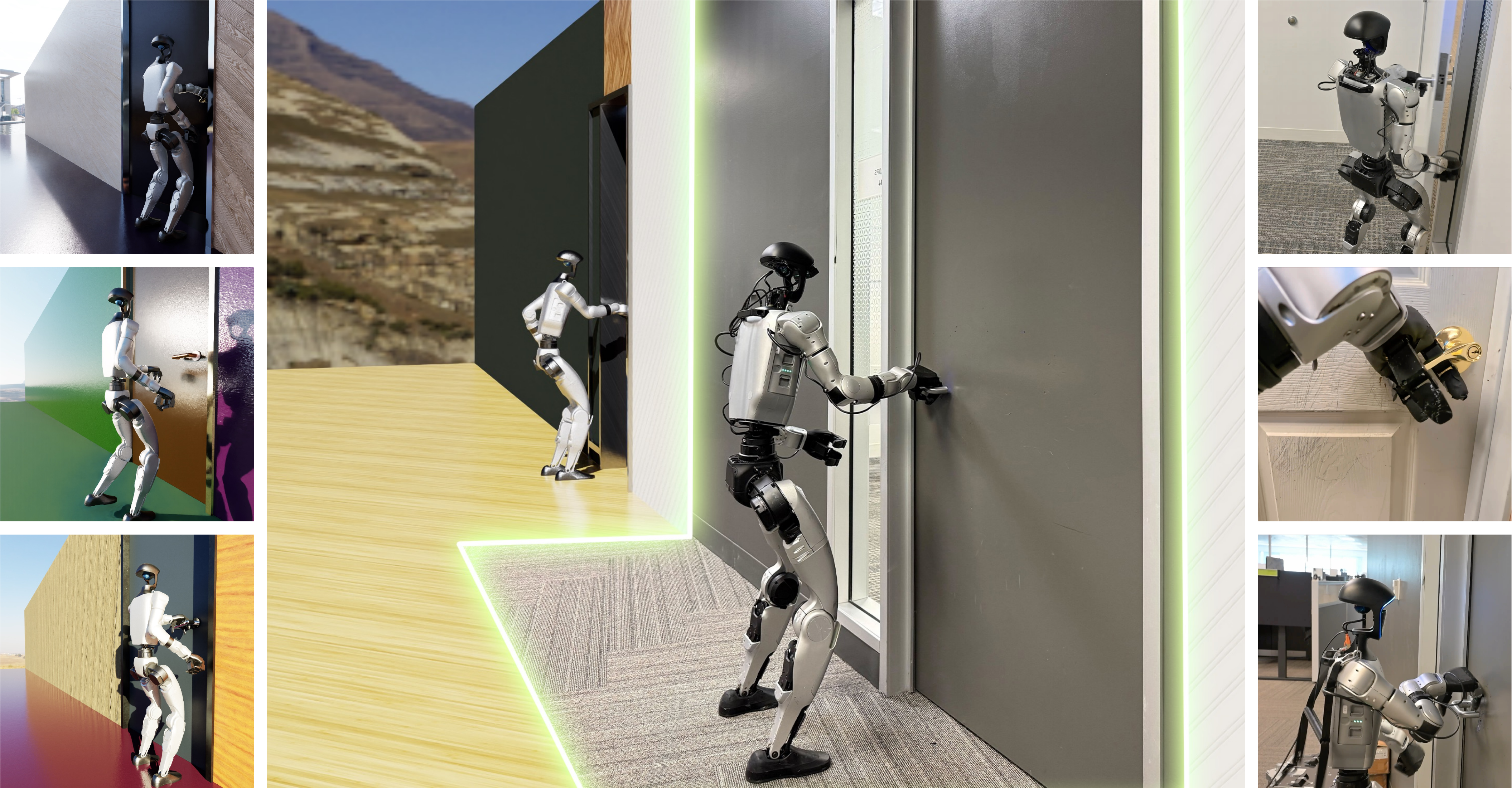}%
  \end{center}
  \vspace{0.3em}

  \begingroup
  \small
  \justifying          
  Figure 1: \textbf{\method{}}, a simulation-trained, RGB-only humanoid loco-manipulation policy, opens diverse, real-world doors.
  \par
  \endgroup

  \vspace{0.5cm}%
}
\makeatother


\maketitle
\newcommand{\tingwu}[1]{\textcolor{blue}{ [TW: \textbf{#1}]}}
\newcommand{\zen}[1]{\textcolor{green}{ [Zen: \textbf{#1}]}}
\newcommand{\ye}[1]{\textcolor{teal}{ [Ye: \textbf{#1}]}}

\begin{abstract}


Recent progress in GPU-accelerated, photorealistic simulation has opened a scalable data-generation path for robot learning, where massive physics and visual randomization allow policies to generalize beyond curated environments. Building on these advances, we develop a teacher-student-bootstrap learning framework for vision-based humanoid loco-manipulation, using articulated-object interaction as a representative high-difficulty benchmark. Our approach introduces a staged-reset exploration strategy that stabilizes long-horizon privileged-policy training, and a GRPO-based fine-tuning procedure that mitigates partial observability and improves closed-loop consistency in sim-to-real RL. Trained entirely on simulation data, the resulting policy achieves robust zero-shot performance across diverse door types and outperforms human teleoperators by up to 31.7\% in task completion time under the same whole-body control stack. This represents the first humanoid sim-to-real policy capable of diverse articulated loco-manipulation using pure RGB perception.

\end{abstract}
\abscontent

\section{Introduction}
\label{sec:intro}

\begin{figure*}[t]
    \centering
    \includegraphics[width=\linewidth]{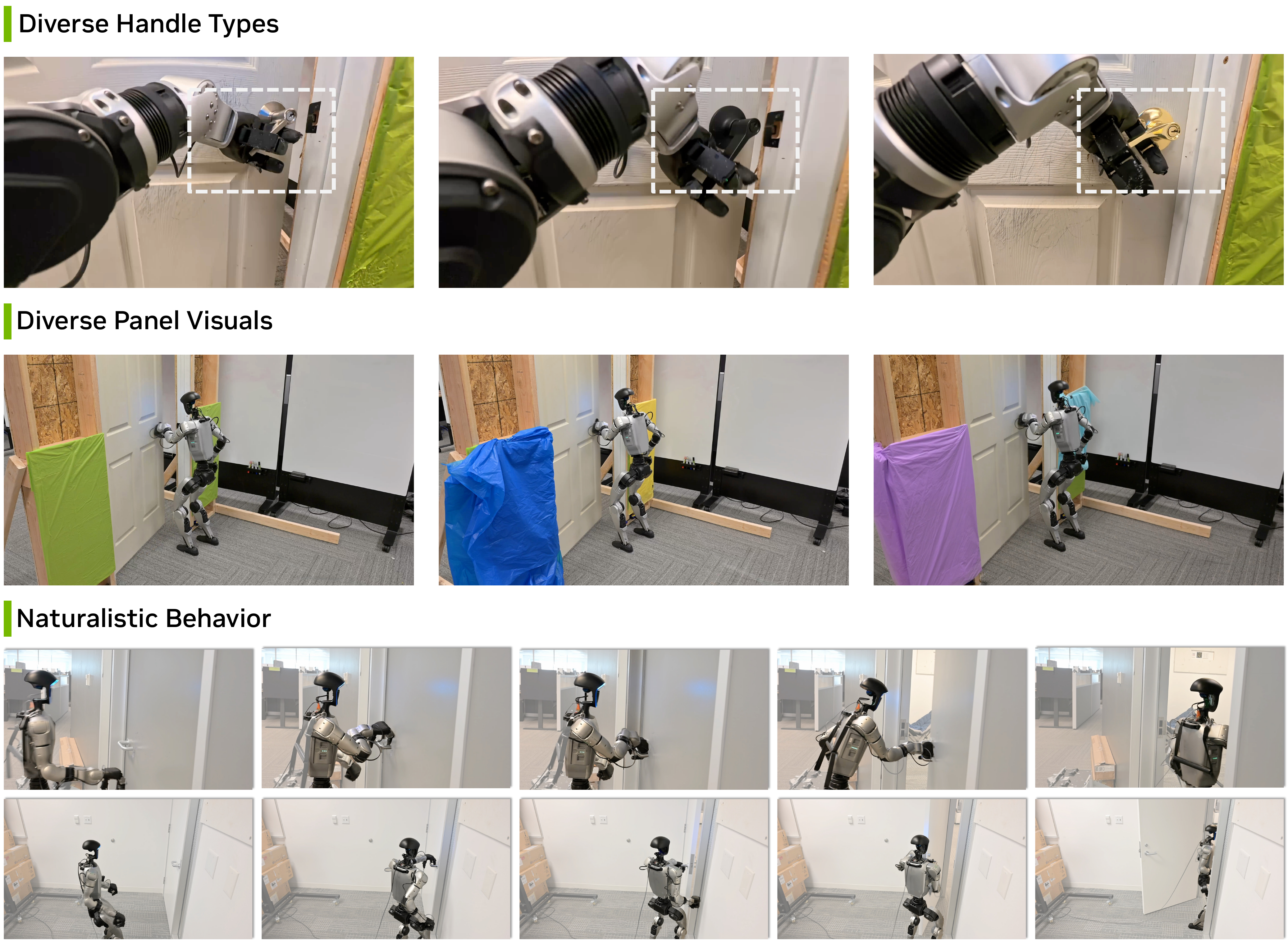}
    \caption{Real-world generalization of \method{}. Top: diverse handle visuals and physical shapes. Middle: diverse wall panel visuals. Bottom: pushing and pulling open doors naturalistically.}
    \label{fig:result}
\end{figure*}

\textit{The reality of robotics is that humanoid kung fu and backflips are solved before they can open doors using only RGB vision.}

Everyday loco-manipulation remains one of the hardest frontiers for humanoid autonomy. Seemingly simple household interactions, such as pulling a drawer, twisting a knob, or unlatching a gate, all require precise perception-action coupling, contact-rich control, and whole-body coordination under uncertainty. Among these tasks, door opening offers a particularly demanding instance: the robot must identify the grasping location from a moving egocentric camera, rotate a spring-loaded handle, track the compliant circular motion of the door panel, and maintain balance under hinge-induced forces. These tightly coupled requirements make door opening a strong stress test for any general-purpose loco-manipulation system.

Our goal in this work is to develop a generalizable learning pipeline for vision-based humanoid loco-manipulation, with door opening serving as a challenging, real-world representative task. Existing systems focusing specifically on doors typically fall short of this broader ambition. Many rely on depth sensing, object-centric features, or hard-coded motion primitives on wheeled platforms~\citep{calvert2025behavior, xiong2024adaptive, weng2025hdmi}. Others simplify contact mechanics or require accurate object localization~\citep{traverse-doors-2025}. DARPA Robotics Challenge-era systems~\citep{oh2017technical} depended heavily on scripting and operator intervention, while more recent teleoperation-centered pipelines~\citep{lee2025stageact} remain brittle. These designs do not produce a scalable solution for the diverse loco-manipulation skills needed in everyday environments.

Recent advances in simulation, hardware, and RL have enabled strong sim-to-real results in locomotion~\citep{zhuang2024humanoid,wang2025beamdojo,long2025learning,ren2025vb,xue2025leverb, ben2025homie}, motion imitation~\citep{liao2025beyondmimic,luo2025sonic,he2025asap}, and dexterous manipulation~\citep{singh2024dextrah,deng2025graspvla,liu2024visual,akkaya2019solving,handa2023dextreme}. However, applying these techniques to loco-manipulation, where perception, balance, contact, and navigation interact, remains under-explored. In this setting, we identify two fundamental challenges for generalizable learning: (i) the algorithm itself must be simple, scalable, and robust to partial observability, capable of producing autonomous policies that coordinate vision and whole-body control (WBC) across diverse tasks. These requirements remain unmet in prior work; and (ii) the visual sim-to-real gap spans a vast space of appearance and physics variation, requiring broad, heterogeneous data rather than a few curated scenes.

To address the first challenge, we introduce a novel, scalable teacher-student-bootstrap learning pipeline. First, a teacher with privileged states (e.g., door pose and articulation state) is trained via reinforcement learning (RL) with stage-conditioned rewards. To improve training efficiency, we introduce an exploration scheme that resets environments from late-stage snapshots, leveraging the recoverability of the simulator. Next, we distill the teacher into an RGB-based student using DAgger~\citep{ross2011dagger}, fusing a vision encoder with proprioception under aggressive visual randomization. Finally, to mitigate the partial observability inherent to vision-only control, we apply GRPO fine-tuning that stabilizes long-horizon behavior and encourages keeping task-relevant regions in view.

To tackle the second challenge, we build a large-scale domain randomization pipeline in IsaacLab~\citep{isaaclab2025} that spans both physics and appearance variation at scale. Physically, we randomize door types, dimensions, hinge damping, latch dynamics, handle placement, and resistive torques. Visually, we randomize materials, lighting, and camera intrinsics/extrinsics. Rather than recreating specific scenes, we intentionally expose the policy to a broad variability envelope, a prerequisite for transferable humanoid loco-manipulation from simulation to the real world.

Across real-world evaluations, the learned policy not only generalizes across diverse articulation mechanisms, appearances, and spatial layouts, but also exceeds human teleoperation in both success rate and efficiency, achieving an 83\% success rate versus 80\% (expert) and 60\% (non-expert), and completing interactions 23.1\%–31.7\% faster using the same whole-body controller, suggesting that our pipeline produces robust, efficient, autonomous loco-manipulation behavior.

\begin{figure}
    \centering
    \includegraphics[width=0.7\linewidth]{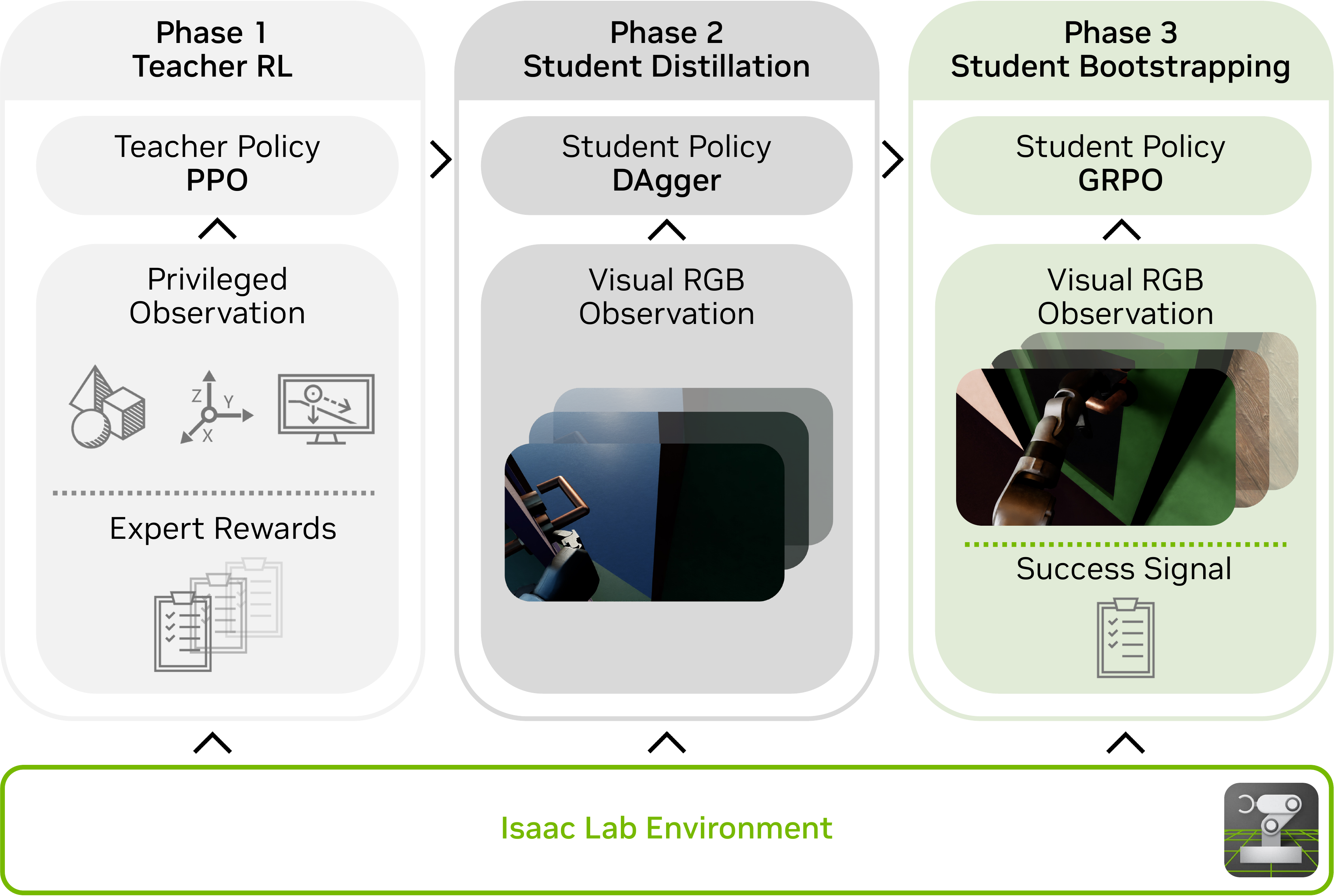}
    \caption{\method{} training pipeline. All phases are done interactively with IsaacLab. In Phase 1, we train a teacher policy with privileged observations. In Phase 2, we distill it into an RGB student policy using DAgger. In Phase 3, we further train the student policy with GRPO using a binary success signal.}
    \label{fig:pipeline}
\end{figure}

To summarize, the main contributions of our work are:

\begin{itemize}
    \item We present one of the first humanoid sim-to-real policy capable of diverse articulated loco-manipulation from pure RGB perception.
    \item We introduce a teacher-student-bootstrap pipeline for whole-body loco-manipulation, including a stage-reset exploration mechanism for stable teacher training and GRPO-based fine-tuning to mitigate the student policy's partial observability.
    \item We provide a physically accurate and visually diverse synthetic-generation pipeline in IsaacLab for humanoid navigation with interactable doors, designed to scale in parallel and distributed RL workflows.
    \item We demonstrate a 31.7\% improvement over human teleoperation using the same whole-body controller, highlighting the potential of photorealistic simulation for scaling vision-based whole-body loco-manipulation learning.
\end{itemize}
\section{RGB Loco-Manipulation via Teacher-Student-Bootstrap}
\label{sec:method}


Here we present \method{}'s three-phase training, building on classical teacher–student distillation. We first outline a visual sim-to-real pipeline for whole-body loco-manipulation, emphasizing two design elements: a multi-stage exploration scheme tailored to long-horizon tasks and a bootstrapped refinement strategy that mitigates partial observability in the student. We then describe a large-scale synthetic generation pipeline in IsaacLab~\citep{isaaclab2025} that produces physically realistic and visually diverse door environments for training and evaluation, treating door opening as a representative loco-manipulation task.

\subsection{Visual RL and Teacher-Student Distillation}

\minisection{Preliminary} Consider a partially observable Markov decision process (POMDP) $\mathcal{P} = (\mathcal{S}, \mathcal{A}, \mathcal{O}, T, \mathcal{R}, \mathcal{O}, \gamma, \rho_0)$, where $\mathcal{S}$ is the state space, $\mathcal{A}$ the action space, $\mathcal{O}$ the observation space, $T(s'|s,a)$ the transition kernel, $\mathcal{R}(s,a)$ the reward, $\mathcal{O}(o|s)$ the observation, $\gamma\in[0,1)$ the discount factor, and $\rho_0$ the initial state distribution. In humanoid whole-body control literature, the policy is responsible for outputting target joint positions, which, in the case of a Unitree G1 robot, includes 29 body joints and 14 hand joints, resulting in an extremely high action space dimension of 33. These joint angles are then tracked by low-level motors using a PD control law. Contrary to quasi-static manipulation literature, the policy needs extremely meticulous reasoning of torque-level dynamics to balance the robot, especially when pushing against a spring-loaded door. The policy also needs to be inferenced consistently at \SI{50}{\hertz}, which requires efficient neural network architectures. We build \method{} on top of a pretrained whole-body controller \citep{ben2025homie} to alleviate the extra burden of handling legged locomotion from scratch.

\minisection{Teacher Policy} The teacher policy $\pi_T(a|s)$ at time $t$ has access to privileged information $o_T\in\mathcal{O}$ that are typically not directly available outside simulation. These include ground-truth robot-root-to-door transform $\xi_\text{RD}$, left-hand- and right-hand-to-door-handle transform $\xi_\text{LD}, \xi_\text{RD}$, net contact wrenches on the 18 hand bodies $\tau_H \in \mathbb{R}^{18 \times 6}$, and root linear velocity $v_R \in \mathbb{R}^3$. While previous works estimate some of these quantities using a hard-coded estimator~\citep{traverse-doors-2025,xiong2024adaptive,calvert2025behavior}, our goal is to eliminate such priors at deployment and maximize generalization with a pure RGB–based student. We train the teacher policy using standard proximal policy optimization (PPO)~\citep{schwarke2025rslrl}, with the exact reward shaping recipe available in Appendix \ref{app:teacher-reward}.

 \minisection{Student Distillation} While the student policy has access to non-privileged proprioception information, such as joint angles $q$, joint velocities $\dot{q}$, and root angular velocities $\dot{\omega}\in\mathbb{R}^3$, its perception of the task relies mostly on the input RGB observation and its temporal context. The image is processed by a vision encoder~\citep{resnet2015}, and the resulting latent is concatenated with the proprioceptive features and passed through a two-layer LSTM (512 units each). A three-layer MLP (512, 256, 128) then maps the recurrent features to target joint angles. The vision encoder is jointly fine-tuned with the policy. The student is interactively distilled using DAgger~\citep{ross2011dagger}, which enables direct supervision on the student's input distribution, compared to behavioral cloning which only covers the teacher distribution.


\subsection{Multi-Stage Whole-Body Loco-Manipulation}

\begin{figure*}
    \centering
    \includegraphics[width=\linewidth]{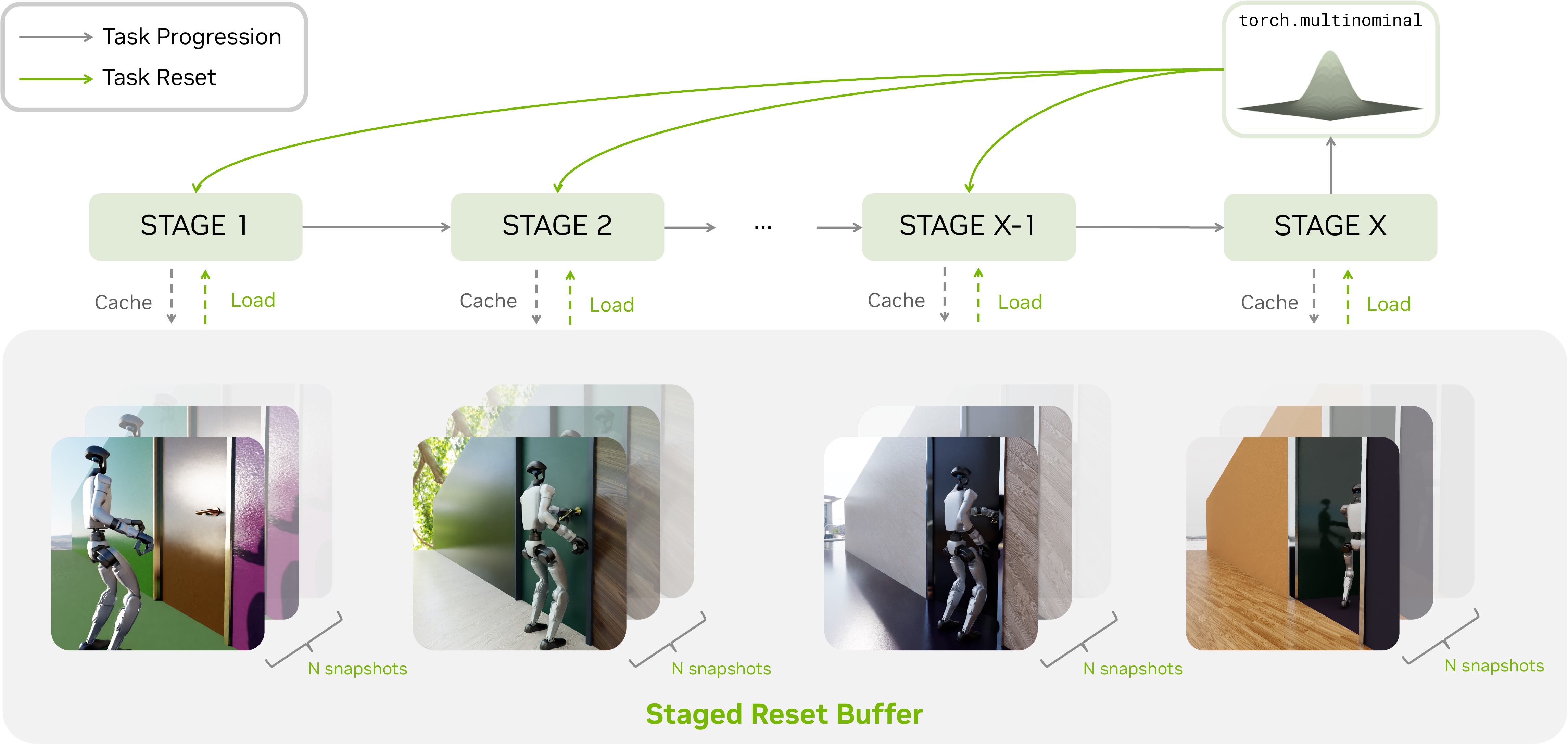}
    \caption{Overview of the staged-reset exploration scheme. When entering a new stage, a snapshot of the simulation is cached into the buffer. When the task resets, the environment is randomly reset to a prior stage by loading from the cache.}
    \label{fig:staged-reset}
\end{figure*}

Here, we present the design of a robust teacher training pipeline for whole-body loco-manipulation tasks. Similar to~\citet{wococo2024}, we design a stage-based reward system to decompose the task into atomic stages, each with its own reward formulations. We inject certain inductive human bias, such as using the door handle and hinge state to distinguish between the approaching, opening, and traversing stages of a successful door-opening operation.

We find that contact-rich tasks that require precise manipulation, such as using articulated doors, present unique challenges in encouraging steady exploration and advancements into later stages. These challenges have not been foreseen in the prior success of RL whole-body control literature. Intuitively, grasping a door handle without the knowledge of carefully rotating it in the correct direction or pairing with precise whole-body movement would incur additional penalties due to the excessive use of motor torque, peaking contact forces, or even the risk of falling over, causing the policy to ``unlearn'' the grasping behavior and keep away from advancing to the next stage.

To improve training efficiency, inspired by~\citet{first_return2021}, we design a simple exploration encouragement scheme leveraging the full recoverability of physical simulators. When an environment proceeds to the next stage, a rolling buffer keeps the recent 100 snapshots of the robot and environment (the door) at that step, which includes the generalized coordinates of all articulated and rigid objects in the scene. Then at reset time, the robot is randomly reset to either the initial stage or one of the middle stages under a nonzero probability. This pipeline is illustrated in Figure \ref{fig:staged-reset}.

To formulate it more formally and clearly see its effect in on-policy RL, consider a long-horizon, multi-stage task, for instance, approaching the door (Stage 1) and opening it (Stage 2). These stages correspond to disjoint subsets $\{S_1,\dots,S_K\} \in S$ connected by narrow transitional regions or bridges $\mathcal{B}_{y,y+1} \in S_y$ that must be traversed to reach the next stage. Because exploration across such bridges has very low probability $p_{\text{bridge}} \ll 1$, policies trained from $\rho_0$ often fail to reach downstream stages early in training, yielding poor long-horizon credit assignment.

To address this, we introduce a \textbf{staged reset law}
\begin{equation}
    \alpha = (\alpha_1, \dots, \alpha_K), \quad \sum_{y=1}^K \alpha_y=1,
\end{equation}
which specifies the fraction of rollouts initialized from each stage's reset distribution $\rho_y$. The resulting initial distribution therefore becomes
\begin{equation}
    \tilde{\rho}_\alpha = \sum_{y=1}^K \alpha_y \rho_y
\end{equation}
with an updated discounted occupancy measure
\begin{equation}
    d_\pi^\alpha(s)=(1-\gamma)\sum_{t=0}^\infty \gamma^t \text{Pr}(s_t=s \vert s_0 \sim \tilde{\rho}_\alpha, \pi),
\end{equation}
where $\text{Pr}$ denotes marginal probability. This shows that the staged-reset scheme reweighs the occupancy measure towards later-stage regions, increasing the frequency and effective magnitude of gradient updates for those states.

\subsection{RL Finetuning for Partial Observability}

In teacher–student policy distillation, a student policy $\pi_S(a|o)$ receives only partial observations $o_t\in\mathcal{O}$, while the teacher policy $\pi_T(a|s)$ has access to privileged observations. Standard behavioral cloning loss alone may not yield optimal performance when the student observation space omits key features due to occlusion. In practice, the student policy often needs to bootstrap on its own rollouts to discover additional strategies that compensate for its partial observability, such as adjusting the robot's position so the manipulated region remains in the camera's field of view.

To enable this self-improvement, we fine-tune the student policy with a Group Relative Policy Optimization (GRPO) \citep{grpo2024} algorithm, which is an actor-only variant of PPO that omits the value function and instead estimates baselines from grouped trajectory scores. Let a batch of $G$ rollouts $\{\tau_i\}_{i=1}^G$ be sampled from the current policy $\pi_S$, each with return $R_i$. We define normalized group-relative advantages
\begin{equation}
    \hat{A}_i = \frac{R_i - \text{mean}(R)}{\text{std}(R)},
\end{equation}
and update $\pi_S$ using the clipped PPO surrogate:
\begin{equation}
    \mathcal{L}_{\text{GRPO}}(\theta) = \Exp{i,t}{\min(r_{i,t}(\theta)\hat{A}_i, \\ \text{clip}(r_{i,t}(\theta), 1-\epsilon, 1+\epsilon)\hat{A}_i},
\end{equation}
where $r_{i,t}(\theta)=\frac{\pi_\theta(a_{i,t}|o_{i,t})}{\pi_\text{old}(a_{i,t}|o_{i,t})}$.

Conceptually, this GRPO fine-tuning phase allows the student to improve beyond imitating the teacher, optimizing its behavior directly under its own partial observations.
Empirically, we observe that such bootstrapping leads the vision-based student to learn compensatory behaviors that the teacher never demonstrated, such as keeping manipulated objects centered in view or adjusting end-effector poses to maintain visibility.
Thus, GRPO acts as a lightweight, stable reinforcement refinement phase that complements behavior cloning, bridging the gap between imitation from privileged demonstrations and robust autonomous performance.

It is worth mentioning that during fine-tuning, we use mainly a binary task success signal, plus simple shaping reward terms such as joint velocity, joint acceleration, and action rate penalty to regularize the humanoid behavior. Therefore, this approach can be a drop-in solution to improve any loco-manipulation task with a base policy of non-zero success rate.

\subsection{Massive-Scale Simulation Randomization}

\begin{figure*}
    \centering
    \includegraphics[width=\linewidth]{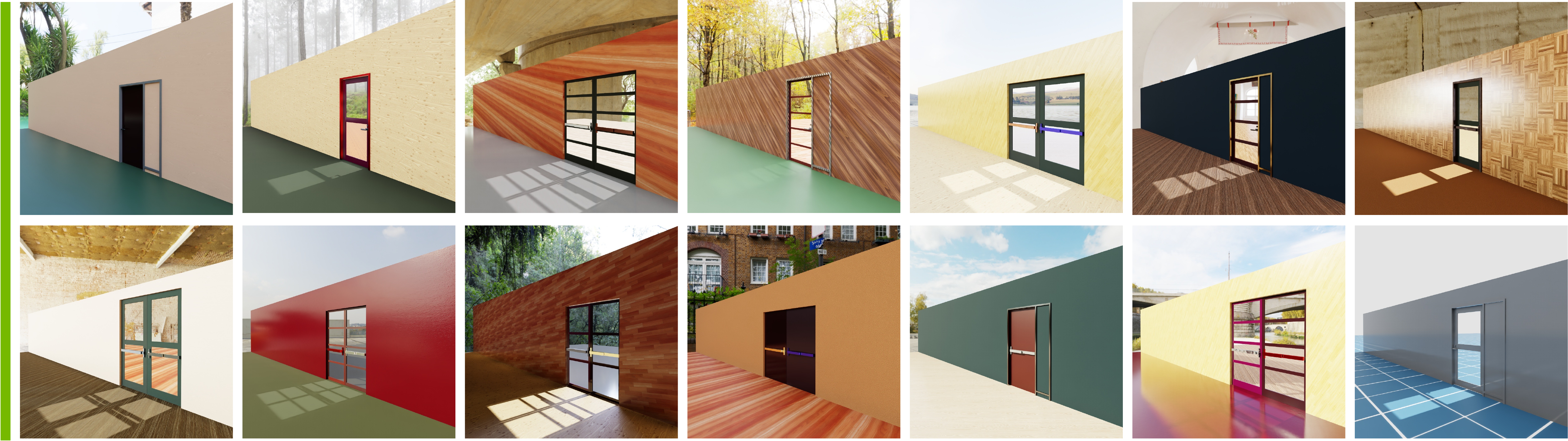}
    \caption{Procedurally generated doors used to train \method{}, covering panel designs, latching mechanisms, lighting, materials, etc. Each parallelized environment is trained on a unique set of door parameters. The last figure shows a door without material.}
    \label{fig:sim-doors}
\end{figure*}

To scale the visual and dynamics diversity of our whole-body loco-manipulation task to an unprecedented level, we design a procedural generation pipeline in IsaacLab that spawns physically and visually diverse yet realistic articulated assets. Compared with prior work such as Infinigen-Sim~\citep{ifinigensim2025}, our IsaacLab-native implementation significantly improves physical realism and enables contact simulation that is both accurate and efficient for parallel RL workflows.

\textit{We emphasize that we do not re-create real-world scenes in simulation; rather, all real-world scenes we evaluate in are unseen during training}. The procedural generation pipeline does not bias towards the specific dimension, physical response, texture, color, or lighting of any real-world location. This contrasts with small-scale behavioral cloning literature~\citep{lee2025stageact} that is confined to be evaluated on the exactly same scene, background, lighting, and time-of-day which the data were originally collected.

\minisection{Physical Variations} We include 5 different door types in the generation pipeline covering commonly seen doors in 3 broad categories: pushing door with rotational handle; pulling door with rotation handle; pushing door with push bar. Similar to~\citet{traverse-doors-2025}, all conceivable aspects of the physical properties of the doors are randomized, such as door dimension, handle location, door-hinge damping, and door-handle resistive torque. Most notably, realistic latching mechanism is used to capture the abrupt change in whole-body dynamics at the moment of opening.

\minisection{Visual Variations} Random textures are drawn from IsaacLab's Physically Based Rendering (PBR) materials and applied to all surfaces. In addition, 5233 dome-light textures are applied to simulate various locations and times of day. To balance rendering quality and performance while training an RL policy in parallel, we use the RTX Real-Time renderer in performance mode, with post-processing effects such as motion blur and auto white balance enabled. Camera extrinsics and intrinsics are aligned and slightly randomized. These settings are essential for re-creating harsh real-world correspondence with the camera being mounted on a legged robot under constant contact switching.
\section{Experiment}
\label{sec:experiment}

In this section, we will establish real-world comparison with human baselines. We will also investigate the effect of varies components in our pipeline, including visual randomization, staged reset, and fine-tuning.

\subsection{Surpassing Human-Teleop Baseline}

\begin{figure}
    \centering
    \includegraphics[width=0.65\linewidth]{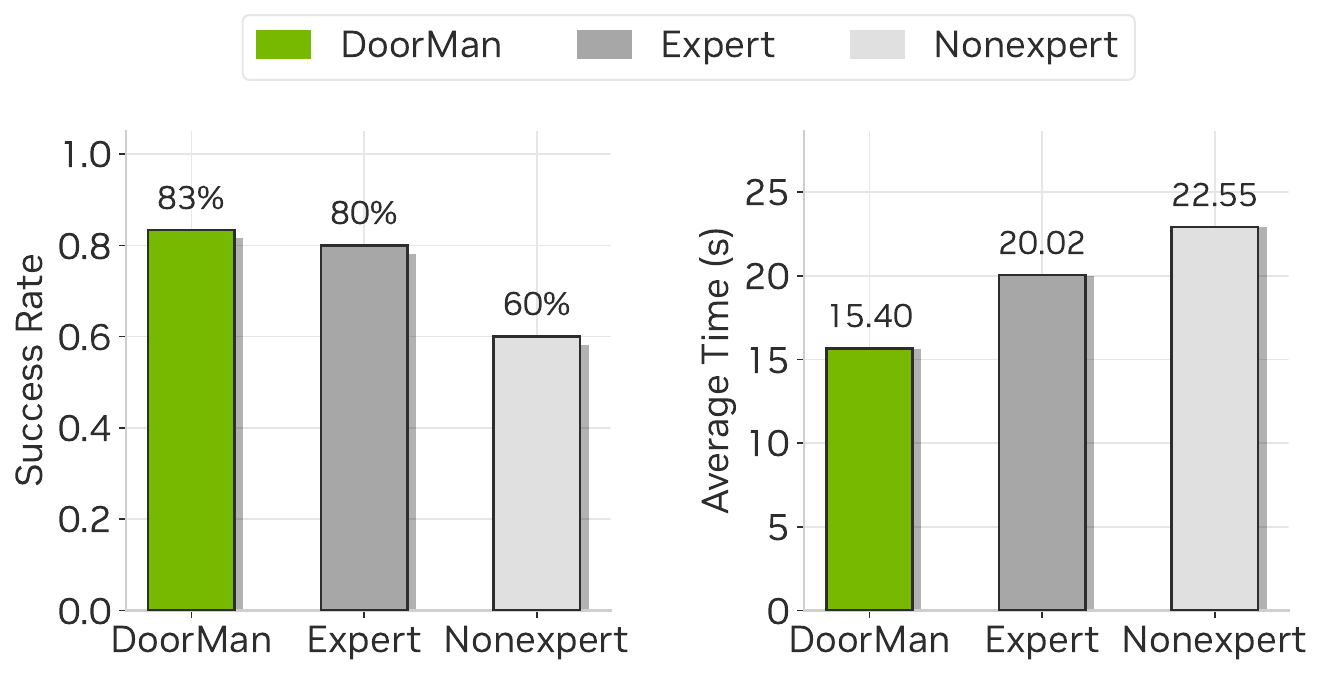}
    \caption{Average performance on all door opening tasks. Left: success rate (the higher the better). Right: task fluency in terms of time taken to complete the door opening task (the lower the better).}
    \label{fig:performance}
\end{figure}

The main question of this work is whether RGB sim-to-real RL can address the long-lasting problem of humanoid door-opening in the wild beyond pure behavioral cloning, whose upper bound is determined by human teleoperation data quality. We hypothesize that the current whole-body teleoperation technology, due to its unintuitive nature, create a gap in both efficiency and success rate compared to direct human operation, and that RGB sim-to-real RL offers better performance.

\minisection{Scenario Setups} We set up evaluation scenarios in both simulation and real world. Three door categories are used:

\begin{itemize}
    \item Rotational handle, opening into the direction of travel \textbf{(push lever)}: the simplest task.
    \item Rotational handle, opening against the direction of travel \textbf{(pull lever)}: requiring skillful manipulation in constrained space and long-horizon behavior.
    \item Push-bar handle, opening into the direction of travel \textbf{(push bar)}: requiring forceful interaction to overcome the spring-loaded hinge.
\end{itemize}

In evaluation, simulation visuals are randomized from textures in a holdout set. Real-world visuals are unseen during training.

In all experiments, the robot is randomly placed to be 1 meter in front of the door and facing the center of the door. Yaw orientation is perturbed by a uniform range of $\pm 0.3$ radians. Success rate and completion time are evaluated at when the robot traverses through the door and reaches a point 1 m beyond the door frame on the opposite side.

\minisection{Human Teleop Baseline} Human teleoperators use the same whole-body controller~\citep{ben2025homie} as the autonomous policy. The teleoperator wears a VR headset with joysticks to control the robot via inverse kinematics, details of which can be found in Appendix \ref{app:teleop}. Teleoperators are classified as \textbf{non-experts} (with less than 1 day of experience teleoperating robots), or \textbf{experts} (with more than 3 months of full-time experience).

Figure \ref{fig:performance} shows that \method{} performs on par in the real world with an expert teleoperator in terms of success rate while being 28\% better than non-experts. In addition, \method{} shines in terms of task fluency, outperforming experts by 23.8\% and non-experts by 31.7\%. Qualitatively, teleoperators often fail to gauge the spring-loaded force of the door handle and the door hinge, or whether the robot is leaning with the appropriate amount to maintain smooth and consistent opening speed. They also often fail to track the revolving path of the swinging door, highlighting a unique challenge in manipulating articulated objects under kinematic constraints, which requires the policy to be aware and compliant of such constraints and maintain whole-body balance at the same time. The feedback information needed for this behavior is beyond the current generation of VR teleoperation, but learnable interactively in simulation.

\subsection{Effect of Photorealistic Visual Randomization}

\textit{How does photorealistic randomization quantitatively affect the generalization on unseen visual features?} We design an ablation study on the visual diversity during training, starting with no visual randomization, where objects are coated in a default gray reflective color, preserving the geometries without textures. We then add 10\% or 100\% of all PBR materials available in randomization, coupled with an additional condition that toggles dome-light randomization to vary lighting and environmental appearance.

Table \ref{tab:vis-rand-ablation} shows that using all available texture and dome light during training yields the best generalization capability in unseen scenarios, with 81-86\% success rates on respective sub-tasks. Removing dome light randomization, and hence the randomization in lighting condition, results in 15-30\% performance drop, with the most significant effect on pulling doors with lever handles, which is the most long-horizon and challenging task. We also observe that with 10\% of all textures available, we can already achieve comparable performance with using 100\%, with only 4-8\% performance drop. Using no visual feature randomization, however, reduces the task success rates to 5-20\%, confirming the effectiveness of texture and lighting randomization in helping sim-to-real policies generalize. We additionally run a setting with only randomizing with solid color materials, paired with dome light randomization, which corresponds to settings in earlier RGB sim-to-real works such as~\citet{tobin2017domain,zhu2018reinforcement}. This setting achieves a commendable success rate of 65.8-70\%. This remaining gap reflects the advantage of modern high-fidelity rendering, which provides richer material and lighting variation and thus stronger visual generalization.

\begin{table*}[t]
\centering
\small
\arrayrulecolor{nvidiagreen}
\setlength{\tabcolsep}{8pt} 

\begin{tabular}{c l c c c c}
\toprule
\textbf{Experiment} & \textbf{Appearance} & \textbf{DL} &
\textbf{Push Lever} &
\textbf{Pull Lever} &
\textbf{Push Bar} \\
\midrule
1 & No Rand.                 & \xmark & 10.8 &  5.0 & 20.0 \\
2 & Solid-color Rand.        & \cmark & 67.5 & 65.8 & 70.0 \\
3 & +10\% Texture Rand.      & \xmark & 58.3 & 50.8 & 76.7 \\
4 & +10\% Texture Rand.      & \cmark & 79.2 & 77.5 & 77.5 \\
5 & +100\% Texture Rand.     & \xmark & 73.3 & 55.8 & 76.7 \\
\rowcolor{bestrow}
\textbf{6} & \textbf{+100\% Texture Rand.} & \textbf{\cmark} &
\textbf{85.8} & \textbf{80.8} & \textbf{85.0} \\
\bottomrule
\end{tabular}

\caption{Success rates (\%) under visual randomization settings.
\emph{Appearance} denotes the type of visual variation:
\emph{Solid-color Rand.} means uniform recoloring without textures;
\emph{+10\%/100\% Texture Rand.} means percentage of texture randomization.
\emph{DL} indicates dome lighting randomization (\cmark{} enabled, \xmark{} disabled).
Each configuration (Experiment~1–6) is evaluated on 120 unseen-door trials.}
\label{tab:vis-rand-ablation}
\end{table*}

\subsection{Performance Boost in GRPO Fine-Tuning}


\begin{figure}[t]
    \centering
    \begin{subfigure}[b]{0.48\linewidth}
        \centering
        \includegraphics[width=\linewidth]{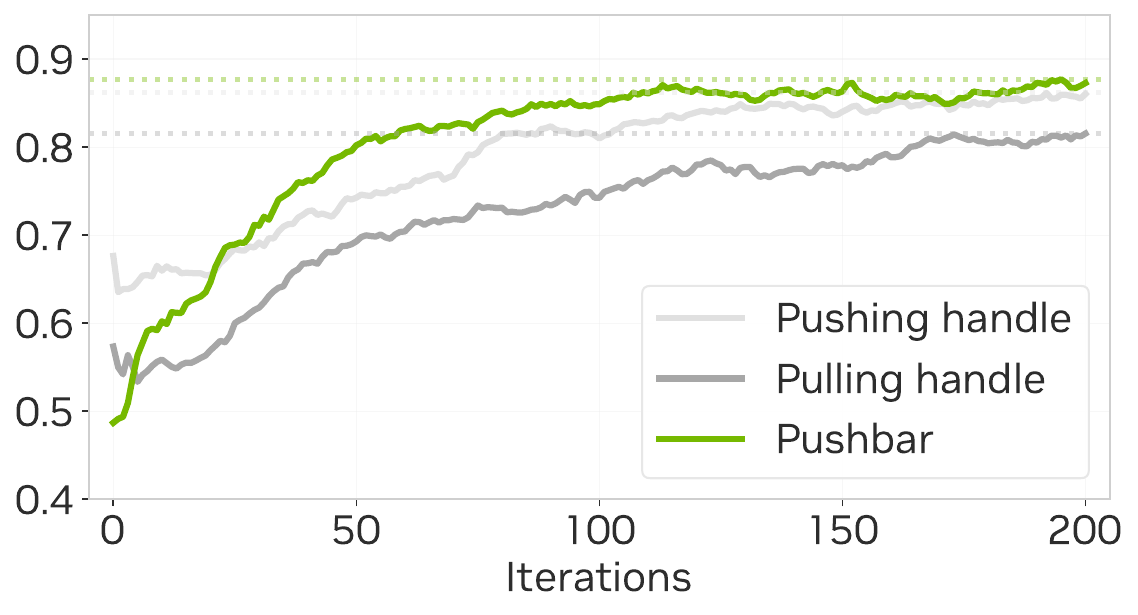}
        \caption{Student policy success rate during GRPO training. 
        Dashed lines show teacher success.}
        \label{fig:reinforce}
    \end{subfigure}
    \hfill
    \begin{subfigure}[b]{0.48\linewidth}
        \centering
        \includegraphics[width=\linewidth]{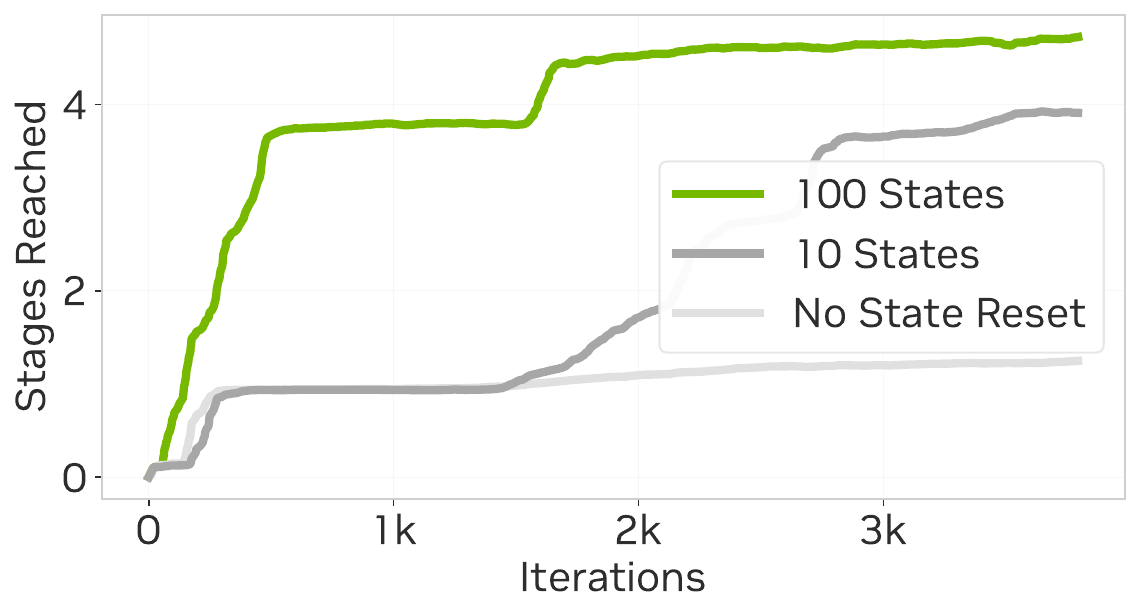}
        \caption{Teacher training progress with reset buffer sizes 0, 10, and 100.}
        \label{fig:staged_reset_ablation}
    \end{subfigure}

    \caption{DoorMan training progress: (a) student GRPO bootstrapping and (b) teacher exploration under different staged-reset buffer sizes.}
    \label{fig:training_progress}
\end{figure}

Figure \ref{fig:reinforce} illustrates the progression of the GRPO bootstrapping phase on 3 door-opening sub-tasks. When teacher policies can consistently achieve 80-90\% success rate, the initial student policy performance stales at 50-70\%, suggesting a non-recoverable observability gap. At the end of the bootstrapping, the student policies achieve 80.8-85.8\% sub-task success rates. Across all open door tasks, the improvement curves exhibit a clear plateau that aligns with the teacher upper bound, indicating that GRPO effectively reduces the gap caused by partial observability and serves as a reliable fine-tuning method for humanoid whole body control.

\subsection{Effect of Staged Reset Exploration}

Finally, we run an ablation study to investigate the effect of staged reset exploration on the stability of teacher training. We use a different set of reset buffer sizes of 0, 10, and 100. A buffer size of 10, for example, stores the ten most recent snapshots of the simulation state when an environment enters a stage. Reward tunings are kept consistent in all trials, and the staged reset exploration is not involved in the reward computation at all. Figure~\ref{fig:staged_reset_ablation} shows that with a buffer size of 100, the teacher rapidly reaches most stages within 500 iterations and all stages by roughly 1700 iterations. With a buffer size of 10, it takes over 4000 iterations for the exploration to finish. The exploration fails when not using the reset buffer, as the policy finds it difficult to enter stage 2 (grasping door handle), which is challenging for incurring additional collision penalties when the policy initially fails to skillfully grasp and rotate the door handle, resulting in ``unlearning'' or avoiding entering into this stage.

\section{Related Work}
\label{sec:relatedworks}

\subsection{Visual Sim-to-Real and Perceptive WBC}





Visual sim-to-real has enabled robust visuomotor control across locomotion~\citep{miki2022learning,zhuang2024humanoid,cheng2023parkour,wang2025beamdojo,long2025learning,ren2025vb,xue2025leverb} and manipulation~\citep{sadeghi2016cad2rl,tobin2017domain,peng2018sim,andrychowicz2020learning,akkaya2019solving,handa2023dextreme,yuan2024learning,jiang2024transic,ze2023visual,hansen2021generalization,huang2022spectrum,singh2024dextrah,deng2025graspvla,liu2024visual,zhu2018reinforcement}. For manipulation, domain randomization has long been used to bridge the reality gap for RGB-trained policies~\citep{sadeghi2016cad2rl,tobin2017domain,peng2018sim}. Teacher–student pipelines further improve generalization by distilling privileged policies into visual ones trained under randomized rendering~\citep{singh2024dextrah,deng2025graspvla}; for example, Dextrah-RGB attains zero-shot dexterous grasping from stereo images~\citep{singh2024dextrah}. For whole-body loco-manipulation, VBC integrates legged locomotion and arm control via hierarchical distillation~\citep{liu2024visual}, and generative pipelines~\citep{yu2024learning} scale visual diversity beyond handcrafted assets. LeVERB~\citep{xue2025leverb} takes advantage of photorealistic demos rendered in IsaacSim for coarse whole-body vision-language tasks. Despite this progress, most prior efforts target isolated arms or decouple locomotion from manipulation, and few demonstrate a vision-only, end-to-end policy that solves tasks demanding whole-body capabilities. We address this gap by learning from RGB and proprioception to produce unified loco-manipulation for door opening, without hand-coded primitives or depth/pose priors.

\subsection{Loco-manipulation}

Loco-manipulation requires a robot to coordinate whole-body motion, balance, perception, and contact-rich manipulation while navigating through its environment. Sim-to-real efforts span modular designs that decouple legs and arms~\citep{liu2024visual,ben2025homie,cheng2024express} and end-to-end policies for whole-body control~\citep{ha2024umi,pan2025roboduet,he2024omnih2o}. Articulated-object interaction (e.g., doors, drawers, and latches) provides a representative instance of loco-manipulation. Prior systems often embed task-specific structure such as rule-based sequences~\citep{oh2017technical,calvert2025behavior}, stagewise controllers with limited grasp synthesis~\citep{lee2025stageact}, or adaptation-heavy pipelines requiring extensive real-world data~\citep{xiong2024adaptive}. Simulation-driven approaches~\citep{urakami2019doorgym,traverse-doors-2025,weng2025hdmi} frequently assume privileged sensing, simplified actuation, or wheeled platforms, limiting their ability to generalize to unstructured environments and whole-body humanoid control.

\subsection{Reinforcement Learning Fine-Tuning}

RL fine-tuning aims to refine a robot policy through additional interaction with the environment. Generalist agents such as RoboCat~\citep{robocat} interleave rollouts, relabeling, and policy updates to gradually expand manipulation skills and robustness~\citep{robocat}. Self-improving visuomotor systems further show that alternating deployment and learning can close the gap between supervised policies and robust real-world controllers~\citep{medals,sime}. Many of these methods use reinforcement learning to adapt an imitation-learned policy to real-world dynamics~\citep{johannink2019residual,ankile2024resip,xiao2025pld}, often requiring substantial additional real-world interaction. Our fine-tuning phase follows this line of work but targets zero-shot sim-to-real transfer for RGB-based humanoid loco-manipulation. After the DAgger distillation phase, we apply GRPO~\citep{grpo2024} in simulation for policy refinement. Combined with extensive domain randomization, this self-refinement yields a policy that keeps the handle in view, maintains stable whole-body balance during traversal, and recovers from off-nominal camera poses.
\section{Conclusion}
\label{sec:conclusion}

In this work, we introduced \method{}, a fully RGB-vision-based learning framework for humanoid loco-manipulation that operates without privileged state estimation. Trained entirely in photorealistic simulation, the resulting policy achieves robust zero-shot performance on articulated-object interaction tasks, including diverse door configurations, and exceeds human teleoperation baselines in both success rate and efficiency. Key ingredients such as staged-reset exploration and GRPO bootstrapping enable stable long-horizon learning and reliable closed-loop behavior under egocentric perception.

This work highlights the potential of large-scale synthetic data and scalable RL pipelines for general humanoid loco-manipulation. Looking forward, we see promising opportunities in reducing dependence on task-specific reward engineering, such as leveraging high-capacity behavioral-cloning teachers, and extending this framework to broader classes of everyday whole-body interactions.

\section*{Acknowledgement}
We thank Jeremy Chimienti, Tri Cao, Jazmin Sanchez, Isabel Zuluaga, Jesse Yang, Caleb Geballe, Beining Han, Chaitanya Chawla, Jason Liu, Tony Tao, Ritvik Singh, Ankur Handa, Arthur Allshire, Guanzhi Wang, Yinzhen Xu, Runyu Ding, Xiaowei Jiang, Yuqi Xie, Jimmy Wu, Avnish Narayan, Kaushil Kundalia, Qi Wang, Scott Reed, Ziang Cao, Fengyuan Hu, Sirui Chen, Chenran Li, Tingwu Wang, Thomas Liao, and Bike Zhang for their help and support during this project.









\clearpage 


\setcitestyle{numbers}
\bibliographystyle{plainnat}
\bibliography{science_template}

\appendix
\newpage
\clearpage
\section*{Appendix}

\section{Teacher Reward Formulations}
\label{app:teacher-reward}

We decompose the door-opening task into six stages: (0) Walk to door, (1) Pre-grasp, (2) Grasp, (3) Open, (4) Swing, and (5) Pass through door. Table~\ref{tab:door-opening-rewards} summarizes the stage-dependent shaping terms used for the teacher policy.

\begin{table*}[t]
\centering
\footnotesize
\setlength{\tabcolsep}{3pt}
\renewcommand{\arraystretch}{1.1}
\begin{tabular}{l l r c}
\toprule
\textbf{Term} & \textbf{Expression} & \textbf{Weight} & \textbf{Stage(s)} \\
\midrule
\multicolumn{4}{l}{\textbf{Termination / Generic penalties}} \\
Termination
  & $\mathbbm{1}_{\{\text{termination}\}}$
  & $-1000.0$ & $0$--$5$ \\

Delta action rate
  & $\|\Delta a_t\|_2^{2}$
  & $-0.01$ & $0$--$5$ \\

DoF velocity
  & $\|\dot{\mathbf{q}}_{\text{upper, non-finger}}\|_2^{2}$
  & $-1.0\times 10^{-3}$ & $0$--$5$ \\

DoF acceleration
  & $\|\ddot{\mathbf{q}}_{\text{upper, non-finger}}\|_2^{2}$
  & $-1.0\times 10^{-5}$ & $0$--$5$ \\

DoF position limits
  & $\sum\max\bigl(0, |\mathbf{q}_i - \mathbf{q}_{\text{limit},i}|\bigr)$
  & $-5.0$ & $0$--$5$ \\

Finger primitive limits
  & $\bigl|\operatorname{clip}(u_{\text{finger}},[l,u]) - u_{\text{finger}}\bigr|$
  & $-1.0$ & $0$--$5$ \\

Humanly DoF limit
  & $\sum\bigl(\operatorname{clip}(\mathbf{q}-\mathbf{q}_{\text{lower}},\text{max}=0) + \operatorname{clip}(\mathbf{q}-\mathbf{q}_{\text{upper}},\text{min}=0)\bigr)$
  & $-1.0$ & $0$--$5$ \\

DoF overspeed
  & $\sum\max\bigl(0, |\dot{\mathbf{q}}_i|-2.0\bigr)^{2}$
  & $-0.1$ & $0$--$5$ \\

Undesired contact
  & $\sum\mathbbm{1}_{\{\|\mathbf{f}_{\text{contact},i}\|>1\}}$
  & $-0.2$ & $0$--$5$ \\

Door frame contact
  & $\sum\|\mathbf{f}_{\text{door frame}}\|_2$
  & $-0.1$ & $0$--$5$ \\

Door panel contact
  & $\sum\|\mathbf{f}_{\text{door panel}}\|_2$
  & $-0.1$ & $0$--$5$ \\

Upright penalty
  & $\bigl\|R_{\text{torso}}[0,0,1]^{\top} - [0,0,1]^{\top}\bigr\|_2^{2}$
  & $-1.0$ & $0$--$5$ \\

HOMIE action limit
  & $\sum\max\bigl(0, |u_{\text{homie},i}|-1.0\bigr)$
  & $-1.0$ & $0$--$5$ \\

\midrule
\multicolumn{4}{l}{\textbf{Stage 0: Walk to door}} \\

Walk to door
  & $\exp\bigl(-\|\mathbf{v}_{\text{robot}}-v_{\text{target}}\hat{\mathbf{d}}_{\text{door}}\|_2^{2}/(2\cdot 0.15^{2})\bigr),\; \sigma=0.15$
  & $5.0$ & $0$ \\

Upper body deviation
  & $\|\mathbf{q}_{\text{upper, non-finger}}-\mathbf{q}_{\text{resting}}\|_1$
  & $-1.0$ & $0$, $5$ \\

Face door
  & $|\text{wrap}_{\pi}(\|\text{axis-angle}(R_{\text{door}})\|_2)|$
  & $-1.0$ & $0$--$2$, $5$ \\

\midrule
\multicolumn{4}{l}{\textbf{Stage 1: Pre-grasp}} \\

Hand-handle orientation
  & $\exp\bigl(-\|\text{wrap}_{\pi}(\|\text{axis-angle}(R_{\text{hand}}R_{\pm 90})\|_2)\|^{2}/(2\cdot 0.6^{2})\bigr)$
  & $3.0$ & $1$--$4$ \\

Pregrasp finger pose
  & $\text{track}(\mathbf{q}_{\text{finger}}, \mathbf{q}_{p0}, \sigma_{\text{pos}}=0.3) + \text{track}(\dot{\mathbf{q}}_{\text{finger}}, 0.6, \sigma_{\text{vel}}=0.2)$
  & $1.5$ & $0$--$1$, $5$ \\

Unused arm deviation
  & $\|\mathbf{q}_{\text{unused arm}}-\mathbf{q}_{\text{rest}}\|_1$
  & $-1.0$ & $1$--$4$ \\

Pre-grasp target distance
  & $\text{track}(\|\mathbf{p}_{\text{hand}}-\mathbf{p}_{\text{pre-grasp}}\|, 0, \sigma=0.2) + \text{track}(\|\mathbf{v}_{\text{hand}}-v_{\text{target}}\hat{\mathbf{d}}\|, 0, \sigma=0.15)$
  & $6.0$ & $1$ \\

Penalty not standing still
  & $\|\mathbf{u}_{\text{HOMIe},[0:3]}\|_2$
  & $-15.0$ & $1$--$3$ \\

\midrule
\multicolumn{4}{l}{\textbf{Stage 2: Grasp}} \\

Grasp finger DoF pose
  & $\text{track}(\mathbf{q}_{\text{finger}}, \mathbf{q}_{p1}, \sigma_{\text{pos}}=0.3) + \text{track}(\dot{\mathbf{q}}_{\text{finger}}, 0.6, \sigma_{\text{vel}}=0.2)$
  & $3.0$ & $2$--$4$ \\

Grasp target distance
  & $\exp\bigl(-\|\mathbf{p}_{\text{hand}}-\mathbf{p}_{\text{grasp}}\|_2^{2}/(2\cdot 0.1^{2})\bigr)$
  & $3.0$ & $2$--$4$ \\

Grasp force
  & $\sum\bigl(-|\mathbf{f}_{\text{palm},y,z}| + f_{\text{palm},x}\bigr)$
  & $0.2$ & $1$--$4$ \\

\midrule
\multicolumn{4}{l}{\textbf{Stage 3: Open door}} \\

Push door handle 
  & $\dot{\theta}_{\text{handle}} + \operatorname{clip}\!\big(\theta_{\text{handle}}, 0, 45^\circ\big) / 45^\circ$ 
  & $6.0$ 
  & $3$ \\

Push door hinge
  & $10\dot{\theta}_{\text{hinge}} + \operatorname{clip}(\theta_{\text{hinge}}, 0, 90^\circ)/90^\circ$
  & $6.0$ & $3$--$4$ \\

Push door force
  & $\operatorname{clip}(\mathbf{f}_{\text{hand},x}, 0, 20)$
  & $0.3$ & $3$ \\

\midrule
\multicolumn{4}{l}{\textbf{Stage 4: Swing door \& Stage 5: Pass through Door}} \\

Don't push door handle
  & $-\dot{\theta}_{\text{handle}} + (45^\circ - \theta_{\text{handle}})/45^\circ$
  & $3.0$ & $4$--$5$ \\

Target root distance
  & $\text{track}(\mathbf{v}_{\text{root}}\cdot\hat{\mathbf{d}}_{\text{target}}, v_{\text{target}}, \sigma=0.2) + \text{track}(\|\mathbf{p}_{\text{root}}-\mathbf{p}_{\text{target}}\|, 0, \sigma=0.2)$
  & $12.0$ & $4$--$5$ \\

Penalty standing still
  & $\exp\bigl(-\|\mathbf{u}_{\text{HOMIE},[0:3]}\|_2^{2}/(2\cdot 0.05^{2})\bigr)$
  & $-1.0$ & $4$ \\

\midrule
\multicolumn{4}{l}{\textbf{Always-on rewards}} \\

Stage progress
  & $\text{stage}_{\text{current}}$
  & $1.0$ & $0$--$5$ \\

Task completion
  & $\mathbbm{1}_{\{\text{complete}\}}$
  & $4.0$ & $0$--$5$ \\

Success save time
  & $\mathbbm{1}_{\{\text{success}\}}\cdot\text{remaining time ratio}$
  & $0.5$ & $0$--$5$ \\

\bottomrule
\end{tabular}
\caption{Reward components for door opening task.  Track($x$, $\mu$, $\sigma$) denotes Gaussian tracking reward $\exp(-(x-\mu)^2/(2\sigma^2))$.}
\label{tab:door-opening-rewards}
\end{table*}

\section{Synthetic Generation Pipeline of Doors}
\label{app:door}

The procedural generation of doors can be divided into two phases. In phase 1, we generate the physical properties of the doors. In phase 2, we apply randomized texture and lighting.

Table \ref{tab:door-rand-properties} summarizes the physical property randomization ranges. We first spawn the geometries for the wall, door panel, push-bar / handle, floor, and latch. Then we add physical joints for the door hinge and handle. The latch is modeled as a mimic joint attached to the joint angle of the handle. Damping, stiffness, and max force are added to the actuators. The door handle actuator is set to have a -5 degrees (upwards) target joint position to simulate the tension of the spring-loaded handle joint even at level position. Additional random features such as key hole, door frame, and other decorations are spawned each at 50\% chance.

We make use of OmniPBR materials and create multiple variants for each by randomizing sub-identifier, texture transform, albedo color, tint color, etc. Every [0.9, 1.1] seconds, a geometry in the scene will have its material randomly drawn. For background dome light texture, we use all publicly available ones in Omniverse, plus an additional 5233 ones from Poly Haven, covering diverse indoor, outdoor, and various times-of-day scenes.

\begin{table}[H]
\centering
\small
\begin{tabular}{@{}lll@{}}
\toprule
\textbf{\textbf{Property}} & \textbf{Range} & \textbf{Unit} \\
\midrule
Panel Width & 0.8-1.1 & m \\
Panel Height & 1.9-2.2 & m \\
Handle Height & 0.85-0.95 & m \\
Handle to Edge Distance & 0.04-0.1 & m \\
Handle Type & \{\makecell[l]{knob, lever, pushbar, handle, flat}\} & \\
Open Handedness & \{left, right\} & \\
Open Direction & \{in, out\} & \\
Weight & 80-120 & kg \\
Hinge Max Force & 20-30 & Nm \\
Hinge Damping & 5-10 & (kg m$^2$) / (s$^2$ $^{\circ}$) \\
Hinge Stiffness & 10-20 & (kg m$^2$) / (s$^2$ $^{\circ}$) \\
Handle Max Force & 1-3 & Nm \\
Handle Damping & 0.1-0.6 & (kg m$^2$) / (s$^2$ $^{\circ}$) \\
Handle Stiffness & 30-50 & (kg m$^2$) / (s$^2$ $^{\circ}$) \\
\bottomrule
\end{tabular}
\caption{Physical property randomization range of doors in IsaacLab.}
\label{tab:door-rand-properties}
\end{table}

\section{Teleoperation Baseline Setup}
\label{app:teleop}

For the experiments in Section~\ref{sec:experiment}, we use a PICO 4 Ultra headset with two handheld controllers for both expert and non-expert teleoperators. The teleoperation interface outputs a command consisting of three upper-body SE(3) poses (head and both wrists), finger joint angles, waist height, and a planar navigation command specifying desired root linear velocity $\mathbf{v} \in \mathbb{R}^2$ and angular velocity $\omega \in \mathbb{R}$ for heading control. We employ the Pinocchio library~\cite{carpentier2019pinocchio} to solve inverse kinematics and map wrist poses to joint-space configurations.

 
\section{Real World Deployment Setup}
\label{app:deploy}

We conduct our experiments on a 29-DoF Unitree G1 humanoid robot, equipped with two 7-DoF 3-finger dexterous hands. Perception is provided by an Intel RealSense D435i camera, without the depth output. Policy inference runs on a desktop workstation with an Intel i9-14900K CPU and an NVIDIA RTX~4090 GPU.

\end{document}